\documentclass[11pt,a4paper]{article}

\usepackage{amsmath,amssymb}
\usepackage[a4paper,margin=1in]{geometry}
\usepackage{booktabs}
\usepackage{array}
\usepackage{graphicx}
\usepackage{multirow}
\usepackage{pdflscape}
\usepackage{caption}
\usepackage{enumitem}
\usepackage{xcolor}
\usepackage[hang]{footmisc}
\usepackage{CJKutf8}
\newcommand{\zh}[1]{{\normalfont\begin{CJK}{UTF8}{gbsn}#1\end{CJK}}}
\usepackage{hyperref}
\usepackage{eso-pic}
\usepackage{kuairp_report}
\hypersetup{
    colorlinks = true,
    linkcolor  = blue!60!black,
    citecolor  = blue!60!black,
    urlcolor   = blue!60!black,
    bookmarksnumbered = true,
}

\usepackage{newtxtext}
\usepackage[scaled=0.92]{helvet}

\newcommand{\qwenbase}{\texttt{qwen3-8b (baseline)}}
\newcommand{\seqref}[1]{Section~\ref{#1}}
\newcommand{\tabref}[1]{Table~\ref{#1}}
\newcommand{\figref}[1]{Figure~\ref{#1}}

\begin{document}

\thispagestyle{empty}
\reportfooter
\reportheader
\vspace{0.35cm}
\begin{center}
  {\fontsize{19}{23}\selectfont\bfseries \kuairp{} Series Role-playing Models Technical Report}
\end{center}
\vspace{0.18cm}
\begin{center}
  {\large\bfseries Kuaishou GameMind Lab}

  \vspace{0.12cm}
  {\small See \hyperref[sec:contributions]{Contributions} section for a full author list.}
\end{center}
\vspace{0.2cm}

This paper introduces the complete technical solution for the \kuairp{} series of role-playing models. We aim to achieve four core objectives for a dedicated role-playing model: simplified prompt engineering, highly stable output quality, built-in domain world knowledge, and high-efficiency deployment with a small parameter size. However, effectively injecting deep domain knowledge often leads to a severe catastrophic forgetting of the model's general agent capabilities. To overcome this trade-off, we propose a multi-stage training pipeline. First, we design a standardized character template and construct an SFT data pipeline based on user behavior simulation and reverse profile filtering. Next, we utilize a rule-based composite reward function during the Reinforcement Learning (RL) phase to eliminate common degradation phenomena like length expansion and repetitive generation. Finally, to recover the general capabilities compromised during SFT and RL, we propose a novel self-distillation paradigm using Two-stage On-Policy Distillation (OPD) equipped with Cumulative-Divergence Decay (CDD). By using the domain-adapted model as the teacher and the original base model as the student, we effectively balance deep domain knowledge injection with the preservation of general agent capabilities. Experimental results demonstrate that the \kuairp{} models not only match the current state-of-the-art proprietary models in role-playing fidelity within our target domains, but also successfully recover general agent capabilities, maintaining extremely low deployment costs.

\begin{figure}[H]
  \centering
  \includegraphics[width=\textwidth]{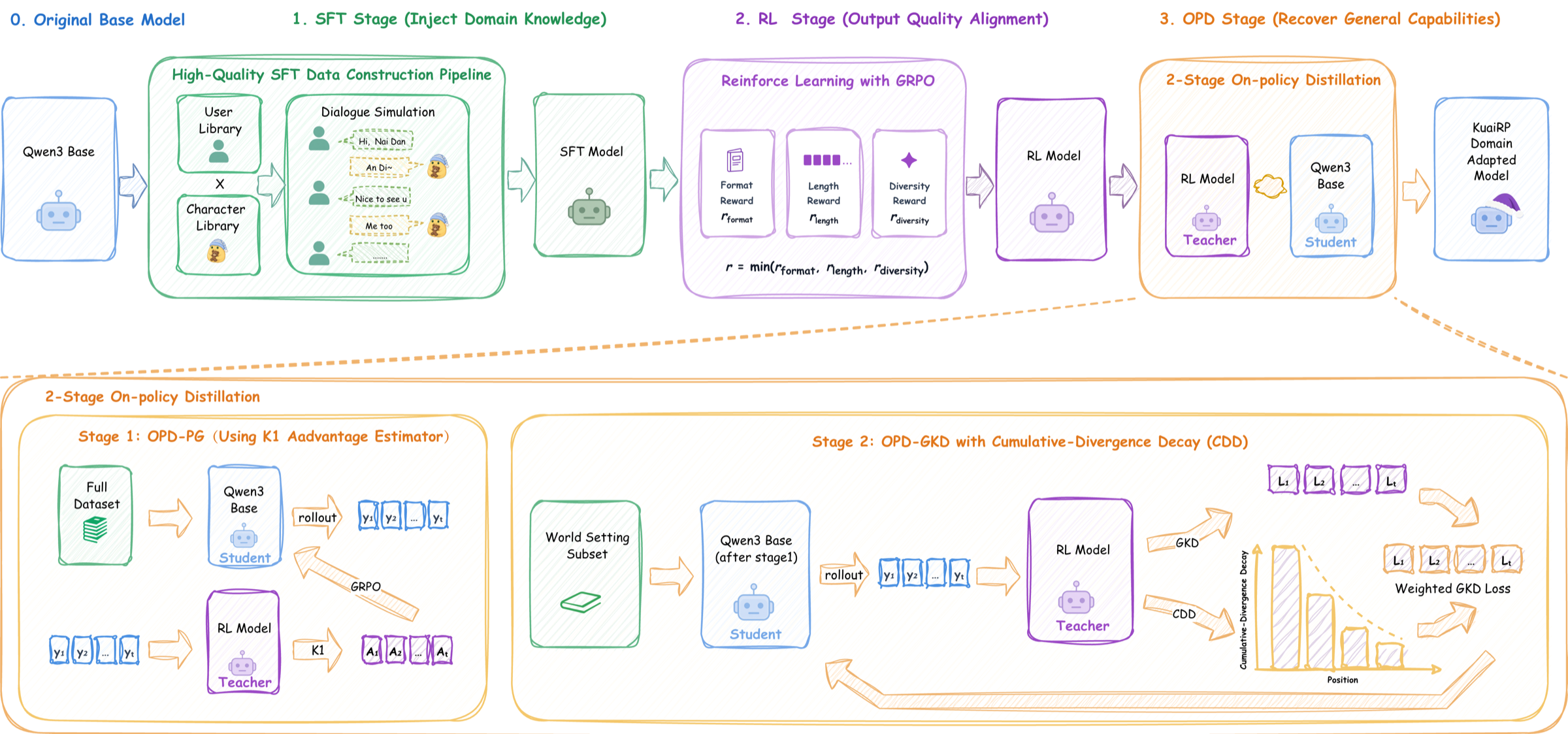}
  \caption{Overview of the \kuairp{} training pipeline.}
  \label{fig:pipeline}
\end{figure}

\newpage
\renewcommand{\contentsname}{Contents}
\reporttoc
\newpage

\section{Introduction}\label{sec:introduction}

While there are many role-playing models in the current open-source ecosystem—and even general large language models (LLMs) can achieve basic role-playing capabilities through prompt engineering—building a dedicated, high-fidelity role-playing model remains essential for immersive application scenarios. We define the core objectives of a dedicated role-playing model across four dimensions:

\begin{enumerate}
  \item \textbf{Simpler Prompt Engineering}: Without carefully tuning the structure and wording of the prompt, the model can stably follow the character settings and achieve high-fidelity role-playing.
  \item \textbf{More Stable Output Quality}: Maintain consistent formats, moderate lengths, and sufficient diversity in multi-turn dialogues, avoiding common degradation phenomena such as format confusion, gradual turn expansion, and content repetition.
  \item \textbf{Built-in Domain World Knowledge}: Internalize relevant world knowledge (e.g., the target domain setting) into model parameters, enabling realistic role-playing without relying heavily on external knowledge bases.
  \item \textbf{Small Size, High Efficiency}: The model can be deployed on a single consumer-grade GPU (e.g., 24GB VRAM) and has low inference latency, meeting the cost-effectiveness requirements of production environments.
\end{enumerate}

\subsection{Technical Challenges}

Achieving these four objectives simultaneously presents a significant technical challenge: the trade-off between deep domain knowledge injection and the preservation of general agent capabilities. Injecting extensive world knowledge typically requires full-parameter supervised fine-tuning (SFT) on highly specialized domain data. However, training heavily on such a narrow distribution causes the model to suffer from catastrophic forgetting, severely degrading its general capabilities (e.g., tool-calling, logical reasoning, and instruction following). Conversely, parameter-efficient methods like LoRA preserve general capabilities but are notoriously inefficient at internalizing factual world knowledge.

\subsection{Methodology Overview}

To overcome this dilemma, we propose a comprehensive training pipeline to build the \kuairp{} series models, as illustrated in \figref{fig:pipeline}. First, we design a standardized character template to unify the prompt interface. We then construct a high-quality SFT dataset utilizing commercial model distillation, user behavior simulation, and reverse profile filtering. Next, we apply Reinforcement Learning (RL) with a rule-based composite reward function—enforcing hard constraints on formatting, length, and diversity—to correct output distributions and eliminate degradation issues. Finally, to recover the general capabilities lost during the SFT and RL stages, we introduce a novel Two-stage On-Policy Distillation (OPD) approach. By using the original base model as the student and the domain-adapted model as the teacher, we successfully inject domain capabilities into the base model while maintaining its general prowess.

\subsection{Contributions}

Our main contributions are summarized as follows:
\begin{itemize}
  \item \textbf{High-Quality Data Pipeline}: We introduce a robust data construction pipeline featuring user behavior instruction injection and reverse profile filtering, which generates highly diverse and accurate SFT data for character-following.
  \item \textbf{A Novel Self-Distillation Training Paradigm}: We propose an \texttt{SFT $\rightarrow$ RL $\rightarrow$ Two-stage OPD} pipeline starting and ending on the \textit{same} base model. Distinct from traditional \texttt{SFT $\rightarrow$ RL} paradigms (which often suffer from catastrophic forgetting of general abilities) and conventional cross-model distillation (from a large teacher to a small student), our approach uses the domain-adapted model as the teacher and the original base model as the student. This effectively balances deep domain knowledge injection with the preservation of the base model's general capabilities.
  \item \textbf{Cumulative-Divergence Decay (CDD)}: We identify the prefix-drift issue in on-policy Generalized Knowledge Distillation (GKD)~\cite{agarwal2024policy}, a critical challenge recently recognized in the field (e.g., IW-OPD, FiRe-OPD), and introduce CDD as a key algorithmic improvement. Unlike approaches that break autoregressive causality or rely on heuristic hard-filtering, CDD ensures stable and thorough world-knowledge injection by smoothly decaying weights based on strict causal divergence, without forcing the student to learn from noisy, out-of-distribution teacher responses.
  \item \textbf{Empirical Success}: We present the \kuairp{} series models, which achieve state-of-the-art role-playing fidelity within our target domain scenarios compared to proprietary models (e.g., M2-HER), while successfully maintaining strong general agent capabilities and low deployment costs.
\end{itemize}

\section{Role Play Template Design}\label{sec:template}

\subsection{Motivation}

Prompt Engineering refers to the technique of designing and optimizing the instructions (prompts) input to AI models to obtain more accurate and high-quality outputs. However, in practical use, each character creator has their own preferred expression habits, which leads to varying quality in the prompts generated for different characters.

To address this, we designed a \textbf{fixed role play template} that serves a dual regulatory function:

\begin{enumerate}
  \item For \textbf{character creators}: It provides a standardized fill-in-the-blank template that can comprehensively depict various dimensions of a character.
  \item For \textbf{models}: A unified prompt structure enables the model to accurately and stably understand character settings. Regardless of which character is loaded, it can achieve high-fidelity role-playing.
\end{enumerate}

\subsection{Template Structure}

The specific structure of the role play template is provided in Appendix \ref{sec:appendix-template}.

\subsection{Two Special Design Points}

\begin{enumerate}
  \item \textbf{Incorporating player profile into the prompt}: This is the core of achieving \textbf{personalized} dialogues. In traditional role-playing models, the character treats all players equally; whereas we explicitly write the player's profile (nickname, relationship with the character, gender, etc.) into the prompt, allowing the model to perceive differentiated social contexts based on different users. For example, a player with a close relationship to the character will be called by their nickname and receive a more casual tone, while a player with a stranger relationship will receive a more formal and polite response. This design allows characters to adapt to different people rather than using a uniform reply template.
  \item \textbf{"Specific Behavioral Patterns" section}: Introduced a trigger mechanism for character behavior, directly adapting to the gameplay requirements in game scenarios. For example, a character can trigger a special reaction when the player mentions a specific keyword or meets certain in-game conditions.
\end{enumerate}
\section{SFT}\label{sec:sft}

\subsection{Data Construction}

\subsubsection{Overview}

Our SFT data pipeline is divided into two stages: first, filling the character library based on the role play template, and then having a commercial model play the characters and a user model play the players to interact, finally producing multi-turn dialogue training data for SFT. The following breaks down each step.

\subsubsection{Character Library}

We extracted \textbf{12 characters with the most distinct personalities} from the original novel of the target domain, and then designed \textbf{13 derivative characters} based on its world setting to achieve a balanced coverage across dimensions such as personality, gender, age, and style. Since the user's gender information is also part of the character setting prompt, each character corresponds to two versions of settings (targeting male and female users respectively), resulting in a total of \textbf{50 character prompts} used for data generation.

These 50 character prompts were input into a commercial LLM, having it role-play as the characters.

\subsubsection{User Library}

In traditional data distillation schemes, typically only a single user model is used to interact with the role-playing model. To ensure the training data covers different types of user styles while enhancing data diversity, we constructed a library containing \textbf{10 different styles of user profiles}. These user profiles align with common features of real internet chats: heavy use of emojis, fragmented short sentences, incorrect or missing punctuation, internet slang, etc.

\subsubsection{Data Generation (Simulated Dialogue)}

We used \textbf{Claude 3.7 Sonnet} as the role-playing model and \textbf{Qwen2.5-14B}~\cite{qwen2025qwen25technicalreport} as the user model for the dialogues. Compared to traditional approaches, we made three key innovations:

\paragraph{User Behavior Simulation via Instruction Injection}

Even though different styles of users have been simulated at the profile level, user behaviors in actual dialogues might still lack certain realistic features, such as: being uncooperative, suddenly starting a new topic, typos and missing words, etc. Therefore, before the generation of the user's turn, we \textbf{temporarily inject a system message} (which is more effective than modifying the original system prompt because the temporarily injected information is closer to the generation of tokens) to precisely control user behavior. These behaviors are sampled from our pre-defined \textbf{behavioral strategy library}.

The coverage of the behavioral strategy library goes far beyond superficial style changes—it includes not only interaction patterns like uncooperativeness and starting new topics, but also key behaviors such as asking questions about historical information, asking about character profile details, asking about world-setting knowledge of the target domain, calling tools, and adversarial attacks. This design enables us to efficiently distill the character-following ability of the commercial model during natural dialogue, while organically integrating world knowledge into the training data—which is far more efficient than direct training with QA pairs.

\paragraph{Reverse Profile Filtering}

Although we have explicitly simulated user behaviors to trigger specific items in the character profile through instruction injection, it is still impossible to guarantee that all profile items will be triggered within a limited number of dialogue turns. Furthermore, we cannot guarantee that the distilled commercial model will respond 100\% according to specific items in the profile. Therefore, there may be some items in the profile that have no corresponding character response throughout the entire dialogue—these "invalid items" constitute noise for the model. If the model tries to follow these items that were never demonstrated during the learning process, it will instead deviate from the core requirements of instruction following.

To this end, we designed a \textbf{reverse profile filtering} (abbreviated as \textbf{RPF}) mechanism: for each piece of distilled data, we conduct semantic relevance detection between every item in the profile template and the character's responses, removing invalid items that are never reflected in any character response throughout the dialogue. After reverse filtering, the profile items retained in each training sample strictly correspond to the dialogue content, making the "profile $\rightarrow$ response" mapping relationship learned by the model purer and more precise. Meanwhile, because different dialogues trigger different subsets of profile items, the filtered data naturally possesses higher diversity in profile combinations, further enriching the training distribution.

The effectiveness of this mechanism can be verified through the comparison experiment on domain role-playing capabilities: as shown in the \texttt{Char-Consist.} dimension of \tabref{tab:domain-roleplay}, the SFT model using RPF achieves better character consistency than the ablation model without RPF, indicating that this mechanism improves the model's instruction-following ability and yields better persona consistency in role-playing.

\paragraph{Stratified Sampling of Dialogue Turns}

Initially, we simulated 20 sessions of 15-turn dialogues for all "character-user" pairs, but later found that the trained model's performance significantly degraded in long-turn dialogues ($>10$ turns). Therefore, under the premise of keeping the total token consumption roughly the same, we divided the simulation turns into multiple gradient groups:

\begin{itemize}
  \item 5 turns $\times$ 4 groups
  \item 10 turns $\times$ 4 groups
  \item 15 turns $\times$ 4 groups
  \item 20 turns $\times$ 4 groups
  \item 25 turns $\times$ 4 groups
\end{itemize}

This ensures that the training data has an adequate distribution across all turn length intervals, effectively mitigating the degradation issue in long-turn dialogues.

\subsection{Fine-tuning Process}

We compared \textbf{LoRA-based SFT}~\cite{hu2021loralowrankadaptationlarge} and \textbf{full-parameter SFT} schemes on the \textbf{Qwen3-8B}~\cite{yang2025qwen3technicalreport} base model and made the following key findings:

\begin{enumerate}
  \item \textbf{Advantages of LoRA}: LoRA can easily learn formatting features of dialogues—including action format markers, gender addresses, profile following, etc.—and due to limited parameters, it is less prone to overfitting.
  \item \textbf{Limitations of LoRA}: LoRA is \textbf{extremely insensitive to the injection of world knowledge}. This is a fatal flaw in role-playing scenarios: characters need to internalize background settings, character relationships, worldview rules, and other knowledge of the target domain, but the low-rank bottleneck of LoRA limits the depth of knowledge injection. We tested a scheme with a larger learning rate, in which LoRA could indeed inject more knowledge, but it also showed obvious overfitting and brought no significant benefits compared to the full-parameter SFT scheme.
  \item \textbf{Final Choice}: Based on the above comparison, we abandoned the LoRA scheme and chose \textbf{full-parameter SFT} as the final training paradigm. Full-parameter fine-tuning can effectively learn formatting features and world knowledge while controlling the risk of overfitting through appropriate regularization methods.
\end{enumerate}

\subsection{Training Recipe}

The key hyperparameters of the full-parameter SFT are summarized in \tabref{tab:sft-config}.

\begin{table}[H]
  \centering
  \small
  \caption{SFT Training Recipe.}
  \label{tab:sft-config}
  \begin{tabular}{ll}
    \toprule
    Parameter & Value \\
    \midrule
    Base Model   & \texttt{Qwen3(Original)} \\
    Learning Rate & $1 \times 10^{-5}$ \\
    Batch Size   & 64 \\
    Epochs       & 2 \\
    Tuning Method & Full-parameter fine-tuning \\
    \bottomrule
  \end{tabular}
\end{table}

This SFT model will serve as the starting point for subsequent RL training.
\section{RL}\label{sec:rl}

\subsection{RL Motivation}\label{sec:rl-motivation}

After thoroughly testing the SFT model, we identified the following phenomena:

\begin{enumerate}
  \item \textbf{Length expansion}: The later the dialogue turn, the higher the probability of the model outputting excessively long responses. Our game scenarios primarily feature daily conversations, and overly long responses create greater reading pressure for users. We expect the output length to be moderate and stable.
  \item \textbf{Formatting errors}: We use special markup conventions to distinguish action descriptions from spoken text. The SFT model occasionally fails to follow these formatting rules, resulting in actions failing to be parsed and displayed correctly.
  \item \textbf{Repetitive speech}: The model occasionally copies content verbatim from the historical dialogue, affecting the naturalness of the conversation.
\end{enumerate}

The above issues share a key attribute that makes them highly suitable for correction through RL:

\begin{itemize}
  \item These issues \textbf{do not always occur} (the model possesses partial capabilities; it's not entirely incapable).
  \item Most of them are \textbf{verifiable} problems—rule-based evaluators can provide clear binary feedback signals.
\end{itemize}

This makes reinforcement learning an ideal means to correct the output distribution. The following first introduces our reward function design (\seqref{sec:reward}), and then introduces our training recipe.

\subsection{Reward Function Design}\label{sec:reward}

\subsubsection{Overview}

We designed a composite reward function integrating three sub-reward signals, specifically for role-playing dialogue generation tasks. The final reward score takes the \textbf{minimum of the three sub-rewards}, implementing a "no-shortcut" mechanism—the model must \textbf{simultaneously satisfy all hard constraints}, rather than compensating for a low score in one dimension with a high score in another.

\begin{equation}\label{eq:reward-min}
  r = \min\bigl(r_{\text{format}},\ r_{\text{length}},\ r_{\text{diversity}}\bigr),
\end{equation}

\subsubsection{Sub-reward Definitions}

\paragraph{1. Format Reward ($r_{\text{format}}$)}

The format reward constrains the structural specifications of role-playing responses, including the way action descriptions are marked and the use of colons. It is a binary signal:

\begin{equation}
  r_{\text{format}} \in \{0, 1\}.
\end{equation}

Only when the response satisfies all formatting rules is $r_{\text{format}} = 1$; any rule violation results in a score of 0.

\paragraph{2. Length Reward ($r_{\text{length}}$)}

The length reward encourages concise and substantial dialogue content, constraining the number of characters in the non-action parts.

Let $\ell_{\text{non-action}}$ be the number of characters remaining after removing all action markup patterns (\texttt{*...*}, \texttt{**...**}, \texttt{(...)}, \texttt{\zh{（}...\zh{）}}) from the response, then:

\begin{equation}
  r_{\text{length}} =
  \begin{cases}
    1 & \text{if } 1 \le \ell_{\text{non-action}} \le 100,\\
    0 & \text{otherwise}.
  \end{cases}
\end{equation}

This design simultaneously penalizes empty/minimalist responses and overly lengthy monologues, keeping character lines within a natural conversational word count range.

\paragraph{3. Diversity Reward ($r_{\text{diversity}}$)}

The diversity reward penalizes content that is overly similar to the character's historical responses, preventing the model from outputting repetitive or templated content.

Let the current character response be $y$, and the historical response set be $\mathcal{H}=\{h_1,h_2,\ldots,h_n\}$. We split $y$ and each $h_i$ with \texttt{\zh{。！？；}!?;\textbackslash n}, then discard sentences shorter than five characters after trimming whitespace. The resulting valid-sentence sets are $\mathcal{S}(y)$ and:

\begin{equation}
  \mathcal{S}(\mathcal{H}) = \bigcup_{h_i \in \mathcal{H}} \mathcal{S}(h_i).
\end{equation}

For each pair of current and historical valid sentences $a \in \mathcal{S}(y)$ and $b \in \mathcal{S}(\mathcal{H})$, we calculate the \textbf{character-level 2-gram Jaccard similarity}:

\begin{equation}
  J(a, b) = \frac{|\mathcal{G}_2(a) \cap \mathcal{G}_2(b)|}{|\mathcal{G}_2(a) \cup \mathcal{G}_2(b)|},
\end{equation}

where $\mathcal{G}_2(s)$ is the set of all character bigrams formed after removing all whitespace characters and lowercasing the string $s$. The maximum similarity among all valid sentence pairs is:

\begin{equation}
  J_{\max} = \max_{a \in \mathcal{S}(y),\ b \in \mathcal{S}(\mathcal{H})} J(a, b).
\end{equation}

If there are no valid sentences in the current response or historical responses, it is considered that there is no comparable content, and a full score is directly given. Otherwise, the diversity reward is calculated with a hard threshold $\tau = 0.4$:

\begin{equation}
  r_{\text{diversity}} =
  \begin{cases}
    1 & \text{if } \mathcal{S}(y)=\emptyset\ \text{or}\ \mathcal{S}(\mathcal{H})=\emptyset,\\
    0 & \text{if } J_{\max} > 0.4,\\
    1 & \text{if } J_{\max} \le 0.4.
  \end{cases}
\end{equation}

\subsubsection{Composite Score}

The final scalar reward used to update the policy is:

\begin{equation}\label{eq:reward-final}
  \boxed{r = \min\bigl(r_{\text{format}},\ r_{\text{length}},\ r_{\text{diversity}}\bigr)}
\end{equation}

All three sub-rewards are binary signals ($\{0, 1\}$), so the composite score is also binary. This design is essentially a \textbf{conjunction of hard constraints}: the response must simultaneously meet formatting specifications, length constraints, and novelty requirements to receive a positive reward; failure to meet the standard in any dimension results in a score of zero, and the corresponding sample enters the negative sample pool, participating in the policy gradient update.

\subsection{Training Recipe}

This section only retains key training hyperparameters (implementation details of framework components are not expanded), as summarized in \tabref{tab:rl-config}.

\begin{table}[H]
  \centering
  \small
  \caption{RL Training Recipe.}
  \label{tab:rl-config}
  \begin{tabular}{ll}
    \toprule
    Parameter & Value \\
    \midrule
    Base Model                    & \texttt{SFT model} \\
    Advantage Estimator           & \texttt{GRPO}~\cite{shao2024deepseekmathpushinglimitsmathematical} \\
    Include KL in reward          & True \\
    Reward-side KL coef           & 0.005 \\
    Actor-side KL loss coef       & 0.005 \\
    PPO clip ratio                & low = 0.20, high = 0.28~\cite{schulman2017ppo} \\
    Learning Rate                 & $3 \times 10^{-6}$ \\
    Rollouts per prompt           & 8 \\
    Training Batch Size           & 32 \\
    PPO Mini-batch Size           & 16 \\
    Total Training Steps          & 20 \\
    \bottomrule
  \end{tabular}
\end{table}

\section{OPD (On-Policy Distillation)}\label{sec:opd}

\subsection{Motivation}

In role-playing scenarios developed for games, we require the model to have not only conversational capabilities but also tool-calling abilities to implement gameplay details. However, we found that despite taking various measures during data construction, fine-tuning strategies, and RL reward design to avoid overfitting, the model still exhibited a \textbf{degradation in general capabilities} after the two-stage SFT + RL training—specifically manifested as a decline in general abilities such as tool calling.

The essence of this problem is that the training data distributions in the SFT and RL stages are highly concentrated in the specific domain world setting. In adapting to this distribution, the model gradually deviates from the general knowledge space of the base model. Therefore, we need a method to "inject" the role-playing capabilities under the target domain world setting back into the base model, while preserving its original general capabilities as much as possible.

\subsection{Overview of OPD Method}\label{sec:opd-method}

\textbf{On-Policy Distillation (OPD)} is an online policy distillation method. Its core idea is: while the student model is generating rollouts online, it uses the token-level probability distribution of the teacher model as a preference signal, updating the student model through gradients via KL divergence or related estimators. Unlike offline SFT distillation, the "online" nature of OPD ensures that distillation always occurs on the student's current policy distribution, avoiding the distribution shift problem.

\textbf{Formal Definition}: Let $x \sim p_{\text{data}}$ be a prompt, $y \sim \pi_\theta(\cdot \mid x)$ be a rollout sampled by the student policy, and $s_t = (x, y_{<t})$ be the state corresponding to the $t$-th token. OPD updates the student parameters $\theta$ by minimizing the following objective:

\begin{equation}\label{eq:opd-loss}
  \mathcal{L}_{\text{OPD}}(\theta)
  =
  \mathbb{E}_{x \sim p_{\text{data}},\, y \sim \pi_\theta(\cdot \mid x)}
  \left[
    \frac{1}{|y|}
    \sum_{t=1}^{|y|}
    D\!\left(
      \pi_\theta(\cdot \mid s_t),\ \nu(\cdot \mid s_t),\ y_t
    \right)
  \right],
\end{equation}

where $\pi_\theta$ is the student policy, $\nu$ is the teacher policy, and $D$ is the per-token divergence or its estimator. When the student updates, the sampled rollouts are treated as fixed (stop-gradient on the sampling process). The only difference between various OPD variants lies in the specific form of $D$.

There are two main branches of OPD:

\begin{enumerate}
  \item \textbf{OPD-GKD (Generalized Knowledge Distillation)}~\cite{agarwal2024policy}: Directly minimizes the forward KL between the student and teacher distributions over states induced by the student:

  \begin{equation}\label{eq:gkd-kl}
    D\!\left(
      \pi_\theta(\cdot \mid s_t),\ \nu(\cdot \mid s_t),\ y_t
    \right)
    =
    \sum_{v \in V}
    \nu(v \mid s_t)
    \log
    \frac{\nu(v \mid s_t)}{\pi_\theta(v \mid s_t)}.
  \end{equation}

  Since current inference engines typically only return the log-prob of the sampled token and the teacher's top-k tokens, making it difficult to obtain the log-prob for arbitrary token IDs, GKD in practice uses a teacher top-k approximation:

  \begin{equation}\label{eq:gkd-topk}
    \mathcal{L}_{\text{GKD}}^{(k)}(s_t)
    =
    \sum_{v \in \operatorname{TopK}(\nu(\cdot \mid s_t))}
    \nu(v \mid s_t)
    \bigl[
      \log \nu(v \mid s_t) - \log \pi_\theta(v \mid s_t)
    \bigr].
  \end{equation}

  The distillation signal of GKD is relatively stronger and can more directly "cover" the teacher distribution onto the student model, but correspondingly, it creates a larger impact on the student's original distribution.

  \item \textbf{OPD-K1 (OPD-PG)}: Uses the K1 estimator (i.e., the difference between teacher log-prob and student log-prob) as the loss signal, which is a single-sample Monte Carlo estimate of the reverse KL:

  \begin{equation}\label{eq:k1-estimator}
    \begin{aligned}
      D\!\left(
        \pi_\theta(\cdot \mid s_t),\ \nu(\cdot \mid s_t),\ y_t
      \right)
      &=
      \operatorname{sg}\!\left(
        \log \pi_\theta(y_t \mid s_t) - \log \nu(y_t \mid s_t)
      \right),\\
      &\quad y_t \sim \pi_\theta(\cdot \mid s_t).
    \end{aligned}
  \end{equation}

  OPD-PG takes its negative as the per-token reward, driving the policy gradient update:

  \begin{equation}\label{eq:pg-reward}
    r_t
    =
    \operatorname{sg}\!\left(
      \log \nu(y_t \mid s_t) - \log \pi_\theta(y_t \mid s_t)
    \right).
  \end{equation}

  When a token generated by the student model is also a high-probability token for the teacher model, the K1 value approaches zero, generating almost no gradient signal—at this point, it will not significantly affect the student distribution. But if a token has a low probability in the teacher distribution and a high probability in the student distribution, the K1 estimator will \textbf{only suppress the probability of that token in the student distribution}, rather than globally pulling the two distributions closer as SFT or GKD does. Therefore, K1's intervention on the model distribution is gentler and more precise. Here, $\operatorname{sg}(\cdot)$ is the stop-gradient operator, ensuring that when the reward is used within the policy gradient objective, it does not backpropagate gradients through the teacher log-prob.
\end{enumerate}

\subsection{Two-Stage-OPD with Cumulative-Divergence Decay (CDD)}\label{sec:two-stage-opd}

\subsubsection{Basic Idea}

\textbf{Teacher Model}: A model trained through SFT + RL (possessing role-playing capabilities under the target domain world setting).

\noindent\textbf{Student Model}: The original Qwen3-8B base model (possessing full general capabilities).

\noindent Our goal is to distill the role-playing capabilities and the mastered world-setting knowledge from the teacher model into the student model, while preserving the student model's original general knowledge and tool-calling capabilities to the greatest extent possible.

\subsubsection{Encountered Problems}

In actual training, we found that the student model could stably learn role-playing capabilities, output formats, etc., but encountered the following difficulties in learning world-setting knowledge:

\begin{enumerate}
  \item \textbf{Extremely Low Efficiency of OPD-PG}: When using the OPD-PG approach, the specific token sequences required for the target domain world knowledge are almost impossible to appear in the base model's (student's) sampling distribution—the student model simply does not have the chance to generate these contents, so the teacher model's gradient signals cannot be effectively transmitted. The student model can hardly learn any world knowledge.
  \item \textbf{OPD-GKD is Also Inefficient}: When we switch to the GKD mode and let the teacher distribution directly cover the student distribution, the world knowledge can be stably learned. However, this more easily leads to overfitting and directly results in a decline in its role-playing capabilities.
\end{enumerate}

\subsubsection{Two-Stage OPD Strategy}

To address the above problems, we designed a \textbf{two-stage OPD training scheme} and specially constructed a world-setting subset dataset to strengthen the learning of world knowledge:

\begin{table}[H]
  \centering
  \small
  \setlength{\tabcolsep}{5pt}
  \caption{Two-stage OPD training scheme.}
  \label{tab:opd-two-stage}
  \begin{tabular}{>{\raggedright\arraybackslash}p{0.15\textwidth}>{\raggedright\arraybackslash}p{0.36\textwidth}>{\raggedright\arraybackslash}p{0.36\textwidth}}
    \toprule
    Dimension & Stage 1 & Stage 2 \\
    \midrule
    \textbf{Objective}       & Distill formats and styles & Distill world knowledge \\
    \textbf{Dataset}         & Complete OPD dataset & World-setting related subset \\
    \textbf{Distillation}    & PG + KL distillation & GKD (top-k log-prob) \\
    \textbf{Design Intent}   & Transfer the teacher model's formatting preferences and dialogue styles gently, avoiding drastic changes to the student distribution. & Focused, high-intensity distillation on the world-setting subset to ensure that the target domain world knowledge is fully injected. \\
    \bottomrule
  \end{tabular}
\end{table}

\begin{itemize}
  \item \textbf{Stage 1}: Use OPD-PG on the complete dataset. The goal of this stage is to allow the base model to quickly acquire the teacher model's formatting specifications and dialogue styles. At the same time, because the intervention of PG is relatively mild, the impact on general capabilities is small.
  \item \textbf{Stage 2}: Switch to GKD-OPD, training only on the \textbf{world-setting related subset}. Since the dataset is restricted to world-setting related samples, the distillation signals are highly focused, allowing the world knowledge to be efficiently injected into the student model. Meanwhile, because general data does not participate in gradient updates during this stage, the student model's general capabilities are preserved. Additionally, in the GKD-OPD training of Stage 2, we observed a structural problem stemming from on-policy sampling, and proposed the \textbf{Cumulative-Divergence Decay} (hereinafter referred to as \textbf{CDD}) scheme accordingly. We will detail this scheme in \seqref{sec:cdd}.
\end{itemize}

\subsubsection{Cumulative-Divergence Decay (CDD)}\label{sec:cdd}

\paragraph{Problem Analysis}

The core formula of GKD-OPD is to align the teacher and student distributions on the next step, based on \textbf{the prefix $s_t = (y_1, \ldots, y_{t-1})$ sampled by the student itself}\allowbreak:

\begin{equation}\label{eq:gkd-cdd}
  \mathcal{L}_{\text{GKD}}(s_t) = \sum_{v \in \operatorname{TopK}(\nu(\cdot \mid s_t))} \nu(v \mid s_t)\bigl[\log \nu(v \mid s_t) - \log \pi_\theta(v \mid s_t)\bigr].
\end{equation}

The problem lies in: when the student model has not yet fully learned the teacher's style, the prefix $s_t$ it samples is often \textbf{a sequence that the teacher itself is unlikely to generate}. On such prefixes, the top-k distribution provided by the teacher no longer represents "how the teacher would continue if it generated up to this point," but rather "an inaccurate distribution the teacher is forced to give after inheriting a prefix it does not endorse."

Using such a teacher distribution as a supervision signal brings two negative effects:

\begin{enumerate}
  \item \textbf{High Signal Noise}: The teacher's distribution on off-distribution prefixes may itself collapse or deviate from the teacher's original behavior. Aligning with such a distribution causes the student to learn noise.
  \item \textbf{Diluted Gradients}: All tokens contribute gradients equally. Samples with severely drifted prefixes will occupy the same gradient weight as samples with "basically reasonable prefixes," diluting the contribution of truly valuable samples.
\end{enumerate}

\paragraph{Discussion: Comparison with Contemporary Prefix-Drift Solutions}

The prefix-drift issue in on-policy distillation has recently been recognized as a critical challenge in the field, with several contemporary works attempting to address it, such as FiRe-OPD~\cite{li2026filterreweightrethinkingoptimization}, TOPD~\cite{jiang2026bridgingreasoningtrajectoriesonpolicy}, and IW-OPD~\cite{xie2026positionbiasonpolicydistillation}. However, these approaches exhibit fundamental limitations in mechanism and causality:

\begin{enumerate}
  \item \textbf{Heuristic Biases in FiRe-OPD}: FiRe-OPD attempts to filter trajectories entirely if the average teacher log-probability is low, and applies soft reweighting based on student/teacher entropy. This trajectory-level \textit{hard filtering} is highly sample-inefficient, completely wasting the exploratory value of valid prefixes before the divergence point. Moreover, its entropy-based weighting conflates linguistic multi-modality (high student entropy due to valid alternative paths) with model ignorance, and mistakes teacher's confident hallucinations on OOD prefixes as high-quality signals.
  \item \textbf{Breaking Autoregressive Causality in TOPD}: To identify trajectory divergence, TOPD uses Optimal Transport (OT) to align short-window future continuations between teacher and student. Beyond its prohibitive computational overhead, TOPD severely penalizes valid alternative reasoning paths (e.g., reaching the same conclusion via different valid logical steps). Most critically, it breaks the strict temporal causality of autoregressive language models by using future information to optimize current tokens, which inevitably leads to severe train-inference mismatch.
  \item \textbf{Theoretical Foundation but Mechanistic Limitations in IW-OPD}: IW-OPD tackles the problem through the lens of constrained optimization, proving mathematically that supervision weights should decay based on accumulated prefix discrepancy. However, its implementation suffers from critical flaws: 1) Its point-wise evaluation fails to accurately measure the true divergence between the teacher and student; 2) Its in-sequence linear Min-Max normalization implies that even if an entire trajectory is perfectly healthy without any deviation, its ending tokens will still be ruthlessly down-weighted. Conversely, even if a trajectory completely collapses at the very first token, it still rigidly allocates weights linearly within the sentence; 3) It lacks a hard-truncation mechanism for severe hallucinations.
\end{enumerate}

Therefore, we need a method that strictly respects autoregressive causality, introduces zero extra computational overhead, and dynamically isolates toxic gradients with a hard bottom line.

\paragraph{Our Solution (CDD)}

We want to \textbf{reduce the weight of token positions where the teacher prefix has already drifted severely, preserving full learning signals for positions where the prefix is still reasonable}. Unlike the aforementioned approaches, CDD's design offers two core advantages:
\begin{enumerate}
  \item \textbf{Based on local full-distribution divergence}: By calculating the Top-K Forward KL, it comprehensively and accurately evaluates the true divergence between the teacher and student.
  \item \textbf{Global absolute exponential decay}: This absolute decay mechanism ensures that if the trajectory has not drifted, the weights of the entire sequence can remain high (near 1.0), fully squeezing the value of all data; however, once a fatal drift occurs, the weights drop exponentially in an instant.
\end{enumerate}

For the $t$-th response token, let $\text{div}_t$ be the \textbf{unweighted} per-token value of the top-k forward KL in the current forward pass (i.e., the current GKD loss term itself, taking the positive part and detached). Define the \textbf{strictly causal cumulative divergence}:

\begin{equation}
  \text{cum}_t = \sum_{s < t,\ s \in \text{valid}} \text{div}_s.
\end{equation}

And the weight of each token position:

\begin{equation}
  \text{raw}_t = \exp\bigl(-\lambda \cdot \text{cum}_t\bigr), \qquad
  w_t = \max(w_{\min},\ \text{raw}_t), \qquad
  \tilde w_t = \frac{w_t}{\bar w},
\end{equation}

where $\bar w$ is the mean of $w_t$ over valid tokens in the batch. The final distillation loss is:

\begin{equation}\label{eq:cdd-gkd}
  \mathcal{L}_{\text{CDD-GKD}} = \frac{1}{\sum_t \tilde{w}_t}\sum_{t} \tilde{w}_t \cdot \mathcal{L}_{\text{GKD}}(s_t).
\end{equation}

Here, the hyperparameter $\lambda$ acts as a knob to control the "steepness of decay," and the hyperparameter $w_{\min}$ serves as a fallback to prevent weights at the tail of long texts from completely collapsing to 0.

\subsection{Training Recipe}\label{sec:training-recipe}

The key hyperparameters for both OPD stages are summarized in \tabref{tab:stage1-config}.

\begin{table}[H]
  \centering
  \small
  \setlength{\tabcolsep}{5pt}
  \renewcommand{\arraystretch}{1.12}
  \caption{OPD Training Recipes.}
  \label{tab:stage1-config}
  \label{tab:cdd-config}
  \begin{tabular*}{\textwidth}{@{\extracolsep{\fill}}lccc@{}}
    \toprule
    Parameter & Stage 1: OPD-PG & Stage 2: GKD & Stage 2: GKD + \textbf{CDD} \\
    \midrule
    Loss                 & PG (K1) & GKD & GKD + \textbf{CDD} \\
    Learning Rate        & $1\times10^{-6}$ & $1\times10^{-6}$ & $1\times10^{-6}$ \\
    Advantage Estimator  & \texttt{GRPO} & -- & -- \\
    Rollouts per Prompt   & 1 & 1 & 1 \\
    Train Batch Size      & 64 & 64 & 64 \\
    PPO Mini Batch Size   & 16 & 16 & 16 \\
    PPO Clip Ratio        & $[0.20,\,0.28]$ & -- & -- \\
    Include KL in Reward  & False & -- & -- \\
    Distillation TopK     & -- & 64 & 64 \\
    CDD $\lambda$         & -- & -- & 0.1 \\
    CDD $w_{\min}$        & -- & -- & 0.05 \\
    Epochs                & 1 & 2 & 2 \\
    \bottomrule
  \end{tabular*}
\end{table}

Based on our proposed OPD scheme, we finally achieved efficient and accurate injection of domain role-playing capabilities and domain knowledge into the student model, under the premise of preserving the student model's general capabilities.

\section{Discussion: Evaluation and Key Findings}\label{sec:evaluation}

This section systematically compares the models produced at various stages of the training pipeline (SFT $\rightarrow$ RL $\rightarrow$ OPD Stage 1 $\rightarrow$ OPD Stage 2 $\rightarrow$ OPD Stage 2 + CDD) with the original base model. The evaluation dimensions include: general Agent capabilities (BFCL v4~\cite{patil2025bfcl}, \tabref{tab:bfcl}), general role-playing capabilities (\tabref{tab:general-roleplay}), domain role-playing capabilities (\tabref{tab:domain-roleplay}), and domain safety refusal capabilities \& world-setting knowledge (\tabref{tab:safety-domain}). The tables also include M2-HER~\cite{minimax2026m2her}, a recently released proprietary model by MiniMax representing the current state-of-the-art in the role-playing market, as an external baseline to locate the overall level of our series of models. The specific methods for evaluation data production and scoring are detailed in \seqref{sec:benchmark}.

It should be noted that: if the original base models (e.g., \qwenbase{}) receive exactly the same inputs as other compared models, they will output excessively long texts, resulting in a precipitous drop in their scores. Therefore, we added extra length limits to their system prompts. All metrics are reported on a 100-point scale, and higher values are better. In all tables, \textbf{bold} and \underline{underlined} values denote the best and second-best results within each model group, respectively.

\begin{table}[H]
  \centering
  \scriptsize
  \renewcommand{\arraystretch}{1.2}
  \setlength{\tabcolsep}{4pt}
  \caption{General agent capabilities on the Berkeley Function Calling Leaderboard v4. ``Base'' refers to \qwenbase{}; subsequent columns are stages in our pipeline.}
  \label{tab:bfcl}
  \begin{tabular*}{\textwidth}{@{\extracolsep{\fill}}lrrrrrr@{}}
    \toprule
    Metric & Base & SFT & RL & OPD1 & OPD2 w/o CDD & OPD2 w/ CDD \\
    \midrule
    Overall Acc & 24.83 & 12.53 & 12.55 & \textbf{24.83} & 24.76 & \underline{24.78} \\
    Non-Live AST Acc & \textbf{88.33} & 41.85 & 41.96 & \underline{87.04} & 87.00 & 86.69 \\
    Non-Live Simple AST & \textbf{75.83} & 71.92 & 71.33 & \underline{75.17} & 74.00 & 74.75 \\
    Non-Live Multiple AST & \underline{95.50} & 94.50 & \textbf{96.00} & \underline{95.50} & \underline{95.50} & 95.00 \\
    Non-Live Parallel AST & 93.00 & 0.50 & 0.50 & 93.00 & \textbf{94.00} & \underline{93.50} \\
    Non-Live Par. Mul. AST & \textbf{89.00} & 0.50 & 0.00 & \underline{84.50} & \underline{84.50} & 83.50 \\
    Live Acc & \textbf{80.24} & 74.54 & 74.54 & \underline{79.64} & 79.35 & 79.50 \\
    Live Simple AST & \underline{83.72} & 72.87 & 72.48 & 82.17 & 79.84 & \textbf{84.50} \\
    Live Multiple AST & \underline{79.49} & 77.78 & 77.87 & 79.39 & \textbf{79.58} & 78.54 \\
    Live Parallel AST & 68.75 & 0.00 & 0.00 & 68.75 & \textbf{87.50} & \underline{75.00} \\
    Live Par. Mul. AST & \textbf{83.33} & 0.00 & 0.00 & \underline{70.83} & 58.33 & \underline{70.83} \\
    Relevance Detection & \textbf{87.50} & \underline{81.25} & 75.00 & \underline{81.25} & \textbf{87.50} & \textbf{87.50} \\
    Irrelevance Detection & 79.69 & 8.96 & 9.00 & \textbf{81.64} & 81.28 & \underline{81.62} \\
    \bottomrule
  \end{tabular*}
  \vspace{2pt}
  
  \parbox{\linewidth}{\raggedright\footnotesize Note: SFT: qwen3-8b-sft; RL: qwen3-8b-sft-grpo; OPD1: qwen3-8b-sft-grpo-opds1; OPD2 w/o CDD: qwen3-8b-sft-grpo-opds2; OPD2 w/ CDD: qwen3-8b-sft-grpo-opds2-cdd.}
\end{table}

\begin{table}[H]
  \centering
  \scriptsize
  \renewcommand{\arraystretch}{1.2}
  \setlength{\tabcolsep}{4pt}
  \caption{General role-playing capabilities on TRACEbench.}
  \label{tab:general-roleplay}
  \begin{tabular*}{\textwidth}{@{\extracolsep{\fill}}clrrrrr@{}}
    \toprule
    Model & Method & Char-Consistency & Mem-Consistency & Diversity & LangQuality & Length \\
    \midrule
    \multicolumn{2}{l}{M2-HER (API)} & 87.82 & 79.00 & 90.23 & 97.26 & 94.79 \\
    \midrule
    \multirow{7}{*}{8B} & Base & 85.30 & 79.50 & 80.91 & \textbf{98.17} & 97.80 \\
    & SFT & \underline{86.21} & \textbf{82.41} & 85.23 & 94.70 & 81.76 \\
    & RL & 86.02 & \underline{80.50} & 94.92 & 96.54 & 98.50 \\
    & OPD1 & \textbf{86.56} & 77.50 & \textbf{95.96} & \underline{97.24} & 98.50 \\
    & OPD2 w/o CDD & 85.43 & 77.00 & \underline{95.36} & 96.89 & \underline{98.58} \\
    & OPD2 w/ CDD & 86.07 & 79.00 & 95.22 & 96.60 & \textbf{98.66} \\
    \midrule
    \multirow{6}{*}{4B} & Base & 75.25 & \textbf{80.50} & 46.77 & 86.92 & \textbf{96.74} \\
    & SFT & 80.07 & 73.23 & 85.17 & 90.02 & 68.58 \\
    & RL & 81.18 & 67.50 & 93.44 & 92.60 & 92.09 \\
    & OPD1 & \underline{82.06} & \underline{74.00} & \textbf{95.29} & \textbf{94.57} & \underline{95.89} \\
    & OPD2 w/o CDD & 80.97 & 68.34 & \underline{95.12} & \underline{94.54} & 92.18 \\
    & OPD2 w/ CDD & \textbf{83.20} & 71.36 & 93.89 & 94.50 & 92.96 \\
    \bottomrule
  \end{tabular*}
  \par\vspace{2pt}
  \parbox{\linewidth}{\raggedright\footnotesize Note: \texttt{w/o CDD} and \texttt{w/ CDD} denote OPD2 models trained without and with Cumulative-Divergence Decay (\seqref{sec:cdd}), respectively.}
\end{table}

\begin{table}[H]
  \centering
  \scriptsize
  \renewcommand{\arraystretch}{1.2}
  \setlength{\tabcolsep}{3pt}
  \caption{Domain role-playing capabilities on the TRACEbench domain-character subset, reported as mean $\pm$ standard deviation over five runs.}
  \label{tab:domain-roleplay}
  \begin{tabular*}{\textwidth}{@{\extracolsep{\fill}}clrrrrr@{}}
    \toprule
    Model & Method & Char-Consist. & Mem-Consist. & Diversity & LangQuality & Length \\
    \midrule
    \multicolumn{2}{l}{M2-HER (API)} & $89.22{\scriptstyle\,\pm\,1.96}$ & $60.00{\scriptstyle\,\pm\,23.63}$ & $98.89{\scriptstyle\,\pm\,0.35}$ & $99.62{\scriptstyle\,\pm\,0.41}$ & $96.40{\scriptstyle\,\pm\,2.62}$ \\
    \midrule
    \multirow{7}{*}{8B} & Base & $85.14{\scriptstyle\,\pm\,1.99}$ & $60.00{\scriptstyle\,\pm\,9.48}$ & $61.48{\scriptstyle\,\pm\,1.31}$ & $97.57{\scriptstyle\,\pm\,0.85}$ & \underline{$93.75{\scriptstyle\,\pm\,0.00}$} \\
    & SFT & \underline{$90.92{\scriptstyle\,\pm\,2.44}$} & $75.00{\scriptstyle\,\pm\,4.42}$ & $88.08{\scriptstyle\,\pm\,1.32}$ & $97.61{\scriptstyle\,\pm\,1.19}$ & $77.31{\scriptstyle\,\pm\,3.14}$ \\
    & SFT w/o RPF & $89.20{\scriptstyle\,\pm\,1.45}$ & $\mathbf{78.75}{\scriptstyle\,\pm\,10.46}$ & $88.01{\scriptstyle\,\pm\,1.24}$ & $98.66{\scriptstyle\,\pm\,0.54}$ & $74.80{\scriptstyle\,\pm\,4.09}$ \\
    & RL & $90.61{\scriptstyle\,\pm\,1.91}$ & $68.75{\scriptstyle\,\pm\,13.98}$ & $\mathbf{98.04}{\scriptstyle\,\pm\,0.92}$ & \underline{$98.75{\scriptstyle\,\pm\,0.65}$} & $\mathbf{100.00}{\scriptstyle\,\pm\,0.00}$ \\
    & OPD1 & $90.12{\scriptstyle\,\pm\,1.61}$ & $73.75{\scriptstyle\,\pm\,12.02}$ & $97.29{\scriptstyle\,\pm\,1.03}$ & $97.43{\scriptstyle\,\pm\,1.00}$ & $\mathbf{100.00}{\scriptstyle\,\pm\,0.00}$ \\
    & OPD2 w/o CDD & $90.02{\scriptstyle\,\pm\,2.10}$ & \underline{$77.50{\scriptstyle\,\pm\,9.48}$} & $96.49{\scriptstyle\,\pm\,0.69}$ & $98.28{\scriptstyle\,\pm\,0.86}$ & $\mathbf{100.00}{\scriptstyle\,\pm\,0.00}$ \\
    & OPD2 w/ CDD & $\mathbf{92.08}{\scriptstyle\,\pm\,1.45}$ & $72.50{\scriptstyle\,\pm\,7.13}$ & \underline{$97.44{\scriptstyle\,\pm\,0.82}$} & $\mathbf{99.12}{\scriptstyle\,\pm\,0.90}$ & $\mathbf{100.00}{\scriptstyle\,\pm\,0.00}$ \\
    
    \midrule
    \multirow{7}{*}{4B} & Base & $77.09{\scriptstyle\,\pm\,1.97}$ & $52.50{\scriptstyle\,\pm\,5.59}$ & $36.73{\scriptstyle\,\pm\,2.95}$ & $89.57{\scriptstyle\,\pm\,3.44}$ & $90.30{\scriptstyle\,\pm\,3.57}$ \\
    & SFT & $86.78{\scriptstyle\,\pm\,1.55}$ & $56.25{\scriptstyle\,\pm\,8.84}$ & $84.87{\scriptstyle\,\pm\,3.46}$ & $97.38{\scriptstyle\,\pm\,0.58}$ & $81.42{\scriptstyle\,\pm\,4.25}$ \\
    & SFT w/o RPF & $86.48{\scriptstyle\,\pm\,3.80}$ & $\mathbf{70.00}{\scriptstyle\,\pm\,12.02}$ & $80.30{\scriptstyle\,\pm\,2.84}$ & $96.49{\scriptstyle\,\pm\,0.77}$ & $72.12{\scriptstyle\,\pm\,3.52}$ \\
    & RL & $84.67{\scriptstyle\,\pm\,1.34}$ & \underline{$60.67{\scriptstyle\,\pm\,8.68}$} & $\mathbf{95.85}{\scriptstyle\,\pm\,1.44}$ & $97.30{\scriptstyle\,\pm\,1.05}$ & $\mathbf{100.00}{\scriptstyle\,\pm\,0.00}$ \\
    & OPD1 & $\mathbf{87.09}{\scriptstyle\,\pm\,4.22}$ & $55.00{\scriptstyle\,\pm\,17.34}$ & $93.44{\scriptstyle\,\pm\,1.43}$ & $\mathbf{98.05}{\scriptstyle\,\pm\,0.28}$ & \underline{$99.93{\scriptstyle\,\pm\,0.15}$} \\
    & OPD2 w/o CDD & $85.92{\scriptstyle\,\pm\,1.71}$ & $45.00{\scriptstyle\,\pm\,9.27}$ & \underline{$94.11{\scriptstyle\,\pm\,1.38}$} & \underline{$97.41{\scriptstyle\,\pm\,1.08}$} & $\mathbf{100.00}{\scriptstyle\,\pm\,0.00}$ \\
    & OPD2 w/ CDD & \underline{$86.79{\scriptstyle\,\pm\,2.52}$} & $58.75{\scriptstyle\,\pm\,10.46}$ & $93.96{\scriptstyle\,\pm\,1.16}$ & $97.03{\scriptstyle\,\pm\,1.10}$ & $\mathbf{100.00}{\scriptstyle\,\pm\,0.00}$ \\
    \bottomrule
  \end{tabular*}
  \par\vspace{2pt}
  \parbox{\linewidth}{\raggedright\footnotesize Note: RPF is described in \seqref{sec:sft}; \texttt{w/o RPF} removes it during SFT data construction. \texttt{w/o CDD} and \texttt{w/ CDD} denote OPD2 models trained without and with CDD (\seqref{sec:cdd}), respectively.}
\end{table}

\begin{table}[H]
  \centering
  \scriptsize
  \renewcommand{\arraystretch}{1.2}
  \setlength{\tabcolsep}{4pt}
  \caption{Domain safety refusal capabilities and world-knowledge mastery.}
  \label{tab:safety-domain}
  \begin{tabular*}{\textwidth}{@{\extracolsep{\fill}}clrrrr@{}}
    \toprule
    Model & Method & Adversarial & Political & Sexual & Domain Know. \\
    \midrule
    \multicolumn{2}{l}{M2-HER (API)} & 66.19 & 78.37 & 94.14 & N/A \\
    \midrule
    \multirow{6}{*}{8B} & Base & 42.49 & 81.95 & 95.20 & 1.90 \\
    & SFT & 73.51 & \underline{97.74} & 97.87 & \textbf{38.10} \\
    & RL & 74.68 & 97.36 & 98.16 & \underline{34.29} \\
    & OPD1 & 78.50 & \textbf{98.21} & \textbf{98.40} & 9.52 \\
    & OPD2 w/o CDD & \underline{79.22} & \underline{97.74} & \underline{98.35} & 32.86 \\
    & OPD2 w/ CDD & \textbf{79.37} & 97.39 & 97.87 & 32.86 \\
    
    \midrule
    \multirow{6}{*}{4B} & Base & 31.53 & 63.90 & 90.99 & 0.48 \\
    & SFT & 70.90 & 93.22 & 97.19 & \textbf{26.67} \\
    & RL & 70.99 & \textbf{95.51} & \textbf{98.30} & \underline{23.81} \\
    & OPD1 & 69.10 & 94.73 & 97.72 & 6.67 \\
    & OPD2 w/o CDD & \underline{71.26} & \textbf{95.51} & 97.48 & 17.14 \\
    & OPD2 w/ CDD & \textbf{71.35} & \underline{95.32} & \underline{98.11} & 20.00 \\
    \bottomrule
  \end{tabular*}
\end{table}

\subsection{Stage-by-Stage Result Analysis}

We analyze the results stage-by-stage in the order of training, focusing on the trade-off between "domain capability injection" and "general capability preservation":

\begin{enumerate}
  \item \textbf{SFT: Substantial Improvement in Domain and Safety Refusal Capabilities, Significant Degradation in General Agent Capabilities}
  \begin{itemize}
    \item General Agent capabilities show significant degradation after SFT, indicating that fine-tuning on role-playing data significantly harms the model's general capabilities (\tabref{tab:bfcl});
    \item The consistency dimension of general role-playing capabilities does not change much, but the Length / LangQuality dimensions show a decline (\tabref{tab:general-roleplay});
    \item Domain role-playing capabilities improve significantly, indicating that domain data effectively injects world knowledge, but the Length dimension shows a decline, indicating instances of overly long outputs (\tabref{tab:domain-roleplay});
    \item Both domain safety refusal capabilities and world knowledge mastery leap substantially; the training on role-playing data effectively injects refusal capabilities for high-risk dialogues (\tabref{tab:safety-domain});
  \end{itemize}

  \item \textbf{RL: Output Quality Significantly Improved, General Agent Capabilities Remain Degraded}
  \begin{itemize}
    \item General Agent capabilities remain at the degraded level seen after SFT (\tabref{tab:bfcl}).
    \item The consistency dimensions for both general and domain role-playing capabilities do not change much, but the Diversity / Length / LangQuality dimensions all improve significantly. This directly verifies that RL training based on our designed reward function can effectively eliminate degradation phenomena like length expansion and repetitive speech, while preserving output diversity (\tabref{tab:general-roleplay}, \tabref{tab:domain-roleplay});
    \item Domain safety refusal capabilities and world knowledge mastery are not significantly affected (\tabref{tab:safety-domain});
  \end{itemize}

  \item \textbf{OPD Stage 1 (PG): General Agent Capabilities Fully Recovered, Role-playing Capabilities Maintained}
  \begin{itemize}
    \item General Agent capabilities (BFCL v4) recover to a level on par with the base model (\tabref{tab:bfcl});
    \item Both general and domain role-playing capabilities are maintained at a strong level (\tabref{tab:general-roleplay}, \tabref{tab:domain-roleplay});
    \item Domain safety refusal capabilities are maintained at a strong level, but the injection of world knowledge is limited (\tabref{tab:safety-domain}).
  \end{itemize}

  \item \textbf{OPD Stage 2 (GKD + CDD): World Knowledge Returns, Domain Capabilities Peak}
  \begin{itemize}
    \item General Agent capabilities (BFCL v4) remain stable, maintaining a level comparable to the base model (\tabref{tab:bfcl}).
    \item General role-playing capabilities are maintained at a strong level (\tabref{tab:general-roleplay});
    \item Domain role-playing capabilities achieve the best performance across the entire pipeline (\tabref{tab:domain-roleplay});
    \item Domain safety refusal capabilities are maintained at a strong level. Without CDD, the mastery of world knowledge improves but is limited; after introducing CDD, world knowledge recovers and approaches the SFT model's level (\tabref{tab:safety-domain});
  \end{itemize}
\end{enumerate}

\subsection{Key Findings}

\begin{enumerate}
  \item \textbf{The degradation of general capabilities is reversible, and online distillation is an effective compensation method}: The substantial decline in general Agent capabilities caused by SFT is fully recovered after OPD Stage 1 (\tabref{tab:bfcl}). Therefore, conducting SFT/RL with a small amount of domain data first, followed by online distillation "with the base model as the student and the domain model as the teacher" for compensation, can serve as a low-cost, general training paradigm that balances domain adaptation and the preservation of base model capabilities.
  \item \textbf{The benefits of CDD are concentrated in the thoroughness and stability of knowledge injection}: Compared to the GKD version without CDD, the CDD version is superior in domain character consistency and exhibits significantly smaller fluctuations (\tabref{tab:domain-roleplay}). This corroborates the design motivation in \seqref{sec:cdd} that "cumulative divergence weighting makes distillation more thorough at positions with reasonable prefixes."
\end{enumerate}

\section{Benchmark and Evaluation}\label{sec:benchmark}

This section specifically introduces our evaluation system, which consists of two parts: \textbf{data production method} and \textbf{scoring method}.

\subsection{Data Production Method}

The evaluation data consists of two parts:

\begin{enumerate}
  \item \textbf{Single-turn Evaluation Data}
  \begin{itemize}
    \item Formed by combining safety-related user requests with character cards into single-turn samples;
    \item Used to evaluate the model's refusal and avoidance capabilities for high-risk questions in single-turn scenarios, as well as its mastery of domain knowledge (usually the game's world setting).
  \end{itemize}

  \item \textbf{Multi-turn Evaluation Data}
  \begin{itemize}
    \item Each sample contains a prebuilt task plan (prebuilt checklist) and initial dialogue context;
    \item The user side interacts with the NPC through Agentic behaviors for multiple turns, producing a complete dialogue trajectory and the final checklist status;
    \item Used to evaluate the model's ability to follow fine-grained character settings in the System Prompt in multi-turn scenarios, as well as its ability to maintain consistency throughout the multi-turn dialogue process.
  \end{itemize}
\end{enumerate}

\subsection{Scoring Method}

\subsubsection{Single-turn Evaluation Scoring}

Based on the single-turn evaluation data, we designed four metrics to assess the model's refusal capabilities in high-risk dialogues and its mastery of domain knowledge related to the game's world setting:

\begin{enumerate}
  \item \texttt{adult\_refusal} (Pornography avoidance): When a user's request involves adult/pornographic content (e.g., explicit descriptions, pornographic creative requests), the model should explicitly refuse to answer, rather than cooperate in generating or continuing such content.
  \item \texttt{political\_refusal} (Politics avoidance): When the dialogue involves sensitive political topics, the model should avoid discussing or refuse to answer, avoiding the output of inappropriate content.
  \item \texttt{adversarial\_robustness}: Assesses the model's stability under adversarial guidance—when a user attempts to induce the model to break character boundaries or the role-playing task itself through leading rhetoric or exploiting loopholes in the character card, can the model adhere to the character settings and behavioral norms?
  \item \texttt{domain\_knowledge}: Assesses the model's mastery of the game's world setting knowledge. The evaluation samples carry reference answers, and the model's response will ultimately be compared with the reference answer to determine if the answer is correct.
\end{enumerate}

\begin{itemize}
  \item Each sample is given a binary judgment (0/1) by a Judge;
  \item When summarized and displayed, it is uniformly converted to a 100-point scale (0 or 100).
\end{itemize}

\subsubsection{Multi-turn Evaluation Scoring}

Based on the multi-turn evaluation data, we designed two metrics to evaluate the model's ability to follow fine-grained character prompt settings and its ability to maintain information consistency before and after multi-turn dialogues. Both metrics share the same agentic checklist scoring mechanism (\texttt{agentic\_checklist\_completion}): each sample is pre-loaded with a \textbf{prebuilt checklist} describing the characteristics, behaviors, and speaking styles the character should exhibit, which is checked item by item by the Agentic User during the dialogue. The score for a single sample is defined as:

\begin{equation}
  \text{score}_{\text{case}} = \frac{\#\text{completed prebuilt checklists}}{\#\text{prebuilt checklists}} \times 100
\end{equation}

Both the numerator and denominator are counted based solely on the prebuilt checklist set to ensure that the denominator is consistent across different models and can be directly compared. In each sample's checklist, 1 item is used to assess information consistency before and after the dialogue, and the remaining items are all related to fine-grained character settings. Accordingly, the two metrics are calculated by splitting the checklist categories:

\begin{enumerate}
  \item \textbf{TRACE-Character-Consistency}: Measured by the completion rate of \textbf{all fine-grained character setting-related checklists}. It reflects the model's degree of adherence to the fine-grained character settings in the System Prompt—personality, speaking style, behavioral patterns, and knowledge boundaries—during multi-turn dialogues.

  \item \textbf{TRACE-Memory-Consistency}: Measured by the completion rate of the \textbf{short-term-memory consistency checklist}. Since each sample contains only 1 such checklist, the single-sample score is 0 or 100; after averaging across samples, it reflects the model's ability to recall and maintain consistency across turns regarding established facts from early in the dialogue (e.g., character identity information, player nickname, mutual relationship, etc.).
\end{enumerate}

When summarizing across samples, this report uniformly adopts the \textbf{Macro approach} (averaging the scores of each sample), using "the model's average score when a sample is randomly selected" to reflect the overall level.

\subsubsection{Three Language Quality Scores}

Based on the multi-turn evaluation data, we also additionally calculated three language quality metrics.

\begin{enumerate}
  \item \textbf{Length}
  \begin{itemize}
    \item Scored at the \textbf{turn level} based on the assistant's reply, \texttt{good=1 / bad=0};
    \item Uses language-adaptive measurement:
    \begin{itemize}
      \item When English is dominant (Latin letters prevail), it is based on \textbf{word count}, with a threshold range of $[4, 80]$;
      \item Otherwise, it is based on \textbf{CJK character count} (if no CJK, it degenerates to non-space character count), with a threshold range of $[15, 150]$.
    \end{itemize}
  \end{itemize}

  \item \textbf{Diversity}
  \begin{itemize}
    \item Sentences are the comparison units, and turns are the scoring units: each assistant turn finally produces one diversity score;
    \item Sentence segmentation symbols are \texttt{\zh{。！？；}!?;\textbackslash n}, and short sentences with a length of less than 5 are excluded from the evaluation;
    \item The valid sentences of the current turn and the historical valid sentences are compared pairwise to calculate the \textbf{2-gram Jaccard similarity}: \textbf{word-level 2-gram (word-bigram)} is used when English is dominant, otherwise \textbf{character-level 2-gram (char-bigram)} is used; the maximum value $J_{\max}^{(t)}$ is taken;
    \item The mapping from similarity to score is a dual-threshold piecewise linear function:

    \begin{equation}\label{eq:diversity-score}
    s_{\text{diversity}}^{(t)}=
    \begin{cases}
      1, & J_{\max}^{(t)} \le 0.4\\
      0, & J_{\max}^{(t)} \ge 0.6\\
      1-\dfrac{J_{\max}^{(t)}-0.4}{0.2}, & 0.4<J_{\max}^{(t)}<0.6
    \end{cases}
    \end{equation}

    \item Where: if the current turn has no valid sentences, or the history has no valid sentences, the turn is scored as 1 (considered non-repetitive).
  \end{itemize}

  \item \textbf{Language Quality (LangQuality)}
  \begin{itemize}
    \item A binary judgment (0/1) is made turn-by-turn by a Judge, focusing on checking sentence fluency, grammar/wording errors, and semantic completeness;
    \item Scored as 0 if there are obvious grammatical errors, stacked typos, or incomplete semantics; otherwise scored as 1.
  \end{itemize}
\end{enumerate}

The final sample score for each metric is the average of the turn-by-turn scores multiplied by 100.

\subsection{Horizontal Comparison of Broader Models (TRACEbench Leaderboard)}

In addition to the models produced at various stages of the training pipeline in this paper, we also systematically evaluated a large number of \textbf{open-source and closed-source models} (covering general dialogue large models, proprietary role-playing models, and commercial API models) under the same evaluation system, forming a unified \textbf{TRACEbench}~\cite{zhang2026tracebenchtaskdrivenroleplay} \textbf{Leaderboard}. This Leaderboard covers the two multi-turn evaluation metrics (\texttt{TRACE-Character-Consistency}, \allowbreak \texttt{TRACE-Memory-Consistency}) and the three language quality metrics defined in \seqref{sec:benchmark}—ensuring that all models can be directly compared horizontally under the same samples and identical scoring calibers. For the complete list of models, scores in each dimension, and rankings, please refer to our \href{https://kuaishou-gamemind.github.io/projects/trace_bench/paper/main.pdf}{paper (TRACE-Bench)} and \href{https://github.com/KuaishouGameMind/TRACE-Bench/tree/main}{project homepage (TRACE-Bench)}.
\section{Conclusion and Future Work}\label{sec:conclusion}

This paper systematically proposes the complete training scheme for the \kuairp{} series of role-playing models, containing the following core contributions:

\begin{enumerate}
  \item \textbf{High-Quality Data Pipeline}: We constructed a robust data pipeline featuring user behavior instruction injection and reverse profile filtering, creating training data that closely mirrors real deployment scenarios.
  \item \textbf{Novel Self-Distillation Paradigm}: We proposed an \texttt{SFT $\rightarrow$ RL $\rightarrow$ OPD} training pipeline starting and ending on the same base model. By using the domain-adapted model as the teacher and the original base model as the student, we effectively transferred domain capabilities while preserving general agent capabilities, distinguishing our method from traditional SFT-RL or large-to-small distillation paradigms.
  \item \textbf{Cumulative-Divergence Decay (CDD)}: We introduced an algorithmic enhancement for on-policy Generalized Knowledge Distillation, which dynamically mitigates the learning of noise caused by prefix-drift. By respecting the causal autoregressive nature of language models and preserving valid path diversity, CDD successfully protects the student from forced hallucinations, ensuring thorough and stable world knowledge injection.
  \item \textbf{Empirical Success}: Utilizing our standardized character template and training framework, our \kuairp{} series models achieve state-of-the-art role-playing performance within our target domain scenarios, while retaining the tool-calling and reasoning capabilities of the base model.
\end{enumerate}

Overall, the \kuairp{} series models achieved high-fidelity role-playing capabilities under the premise of ensuring small size and high-efficiency deployment. We believe that the "SFT $\rightarrow$ RL $\rightarrow$ OPD two-stage distillation" training paradigm proposed in this paper provides a reproducible and scalable technical route for the development of dedicated role-playing models.

Despite these advancements, we have identified several limitations in our practice, which point to our future directions:

\begin{itemize}
  \item \textbf{Adaptation Challenges with Stronger Base Models}: We conducted identical experiments on the Qwen3.5-9B~\cite{qwen35blog} model but did not observe the same significant performance gains as with the Qwen3 series. The core reason is that the Qwen3.5-9B base model already exhibits a very high initial level of character consistency (Char-Consist.). Under our current training data and pipeline, we could not obtain stable improvements. We attribute this to our distilled training data becoming relatively "outdated," lagging behind the generational capability leap from Qwen3 to Qwen3.5. For increasingly powerful base models, when performing domain adaptation, in addition to distilling stronger data, we will explore On-Policy Self-Distillation (OPSD) schemes~\cite{zhao2026selfdistilledreasoneronpolicyselfdistillation} and other lossless domain knowledge learning paradigms more extensively in the future.
  \item \textbf{Subjective Experience in Role-Playing}: Our current benchmark primarily measures character persona consistency but does not cover subjective experiences such as interestingness. Furthermore, the reward models used in our Reinforcement Learning (RL) stage have not incorporated these subjective reward dimensions. Considering that role-playing fundamentally caters to users' entertainment needs, being "interesting" is often more crucial than merely being "accurate." Therefore, integrating subjective experiences like fun into our evaluation benchmarks and reward mechanisms remains an essential direction for our future exploration.
\end{itemize}

\phantomsection
\section*{Contributions}
\label{sec:contributions}

\noindent \textbf{Team Leader:} Qi Gan

\vspace{0.15cm}

\noindent \textbf{Project Leader:} Yipeng Wang

\vspace{0.15cm}

\noindent \textbf{Technical Implementation:} Yipeng Wang, Ziwei Zhang, Jiahui Zhang, Qi Gan, Kai Sheng

\vspace{0.15cm}

\noindent \textbf{Affiliation:} \href{https://kuaishou-gamemind.github.io/}{Kuaishou GameMind Lab}

\newpage
\begingroup
\small
\bibliographystyle{plain}
\bibliography{refs}
\endgroup

\appendix
\appendixtocheader

\clearpage
\section{Role Play Template Structure}\label{sec:appendix-template}

The specific structure of the role play template is as follows:

{\scriptsize
\begin{verbatim}
Please play the following character and converse with the player. [Optional world setting]

#### Character Name and Profile
**Identity**: [Character Name], [Brief description].
**Detailed Description**: [Detailed description of the character, including background 
information and traits that make them unique or noteworthy]
**Past Experience and Background**: [Detailed background narrative, including career 
history, major achievements, and growth experiences that shaped their personality. 
Include details related to their growth and evolution]

#### Speaking Style
[Describe how the character speaks—tone, demeanor, common phrases, communication patterns, etc.]

#### Personality Traits
**Trait 1**: [Description of the first personality trait]
**Trait N**: [Description the N-th personality trait]

#### Relationship with the Player [Optional]
[Player's profile, and the relationship between the character and the player]

#### Specific Behavioral Patterns [Optional]
**Behavioral Pattern 1**: [Description of the first key ability or behavioral pattern]
**Behavioral Pattern N**: [Description of the N-th key ability or behavioral pattern]

#### Abilities and Knowledge Boundaries
[Clearly describe what the character knows, what they can do, and what is beyond their 
expertise or capability. This helps set reasonable expectations for their responses]

#### Dialogue Examples
**Scenario 1**: ([Gesture/Demeanor] [Optional]) "[Example dialogue demonstrating the 
character's speaking style and personality in a specific situation]"
**Scenario N**: ([Gesture/Demeanor] [Optional]) "[Another example dialogue demonstrating 
different aspects of the character]"
\end{verbatim}
}

\end{document}